\documentclass[runningheads]{llncs}
\usepackage[T1]{fontenc}
\usepackage{graphicx,verbatim}
\usepackage{hyperref}

\usepackage{booktabs}
\usepackage{multirow}
\usepackage{graphicx}
\usepackage{comment}
\usepackage{amsmath}
\usepackage{xcolor}
\usepackage{enumitem}
\usepackage{parskip}
\usepackage{orcidlink}

\usepackage{array}

\usepackage[numbers]{natbib}
\usepackage{color}

\usepackage{amsfonts}

\begin{document}
\titlerunning{Aggregated Aleatoric Uncertainty Fails to Capture Presence Ambiguity} 

\title{Beyond Boundary Noise: Aggregated Aleatoric Uncertainty Fails to Capture Presence Ambiguity in 3D Lung Nodule Segmentation}
%

\author{
Simon Baur\inst{1}\orcidlink{0009-0009-4307-3078}, 
Arne Schernich\inst{1}\orcidlink{0009-0003-3722-6564},
Ekin Böke\inst{1}, 
Wojciech Samek \inst{1,2,3}\orcidlink{0000-0002-6283-3265},
Jackie Ma\inst{1}\orcidlink{0000-0002-2268-1690}
}

\authorrunning{Simon Baur et al.}

\institute{Fraunhofer Heinrich-Hertz-Institut, 10587 Berlin, Germany\\
\email{\{simon.baur, wojciech.samek, jackie.ma\}@hhi.fraunhofer.de}\\
\and
Technische Universität Berlin, 10623 Berlin, Germany\\
\and 
The Berlin Institute for the Foundations of Learning and Data (BIFOLD), 10587 Berlin, Germany\\
}
  
\maketitle              
\begin{abstract}
Uncertainty estimation is critical for the safe clinical deployment of deep
learning in medical image segmentation, with aleatoric uncertainty
theoretically designed to capture irreducible data ambiguity. However,
whether entropy-based measures reflect clinically meaningful ambiguity, i.e.
case-level disagreement about whether a pathology is present at all, remains
poorly understood. Contrary to most prior work, which focused on pixel-wise
boundary disagreement, we systematically evaluate how well aleatoric
uncertainty captures presence ambiguity. Our evaluation spans 3D lung nodule
segmentation across four architectures with Monte Carlo dropout and deep
ensembles, on LIDC-IDRI and an external validation cohort (LNDb). We find
that entropy-based uncertainty maps align with boundary noise and minor
drawing variation but carry insufficient discriminative signal for presence
ambiguity. In contrast, a lightweight supervised ambiguity head trained on
frozen segmentation features substantially outperforms all
entropy-aggregation-based baselines across architectures, metrics, and both
cohorts, and matches or exceeds methods that explicitly model ambiguity
under disagreement supervision (Probabilistic U-Net, Annotator-Confusion
3D-UNet). A qualitative feature-space analysis shows that presence ambiguity
is already encoded in the frozen encoder features of pixel-wise-trained
networks, only to be discarded by the segmentation output and its entropy
aggregation. Our findings expose a fundamental mismatch between the
theoretical promise of aleatoric uncertainty and its practical behavior, and
suggest that practitioners should not rely on entropy-based uncertainty as a
proxy for clinical ambiguity in safety-critical applications.


\end{abstract}

\section{Introduction}
\begin{figure}[!h]
\includegraphics[width=\textwidth]{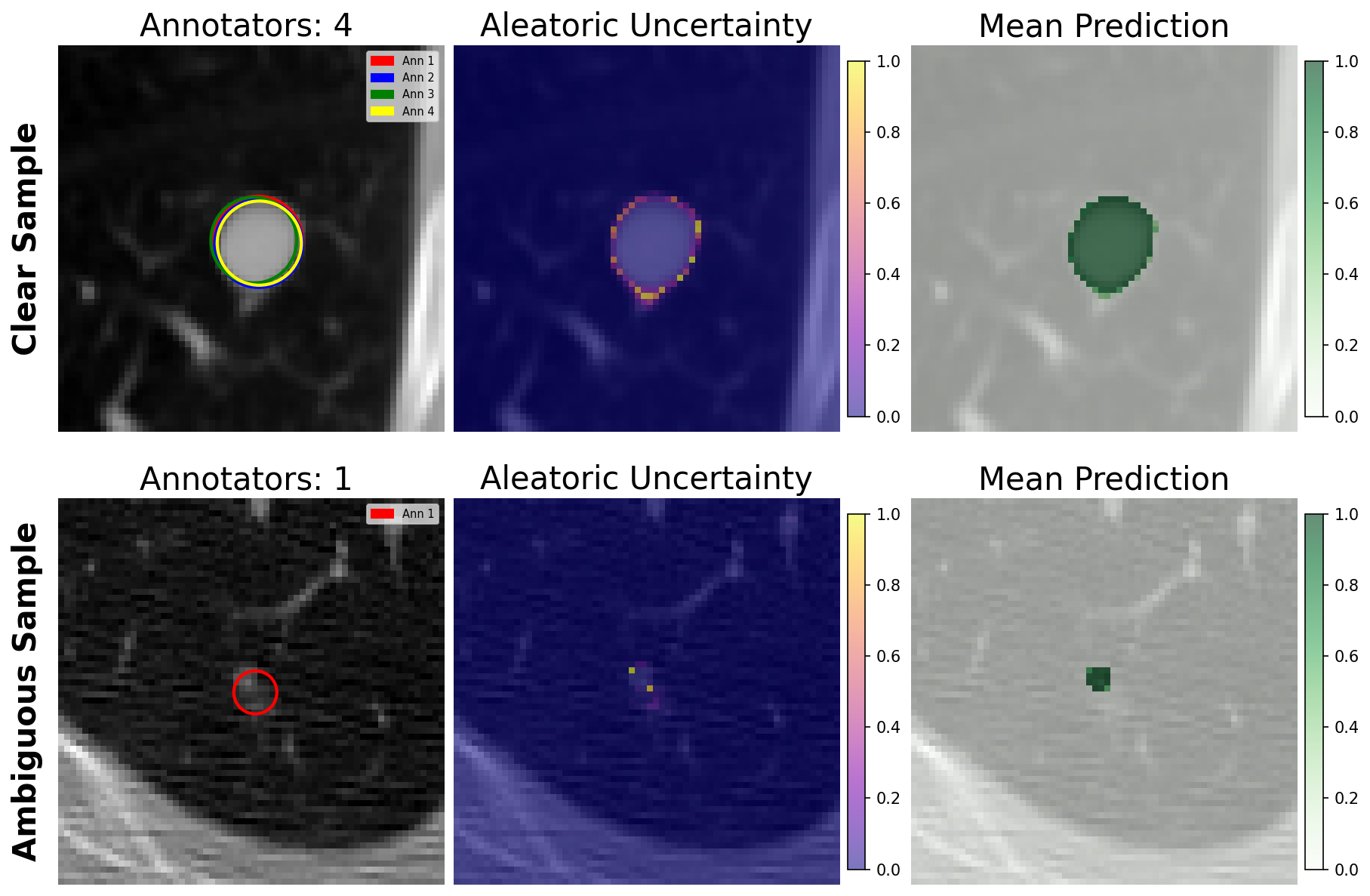}
\caption{Visualization of a non-ambiguous (top row) and ambiguous (bottom row) lung nodule patch from the LIDC-IDRI dataset. Each column shows: the CT image with annotator segmentations overlaid as colored circles (left), the aleatoric uncertainty map derived from MC-Dropout (center), and the mean predicted segmentation mask (right). We can clearly see that the ambiguous sample (where annotators disagree on whether a structure is present at all) produces virtually no aleatoric uncertainty signal.} \label{fig:sample}
\end{figure}

Machine-assisted medical decision making increasingly supports clinical workflows such as diagnosis, treatment planning, and disease monitoring \cite{thomas2025artificial}, where erroneous or overconfident predictions can directly affect patient outcomes \cite{li2025reducing}. Models must therefore provide not only accurate predictions but also reliable uncertainty estimates \cite{begoli2019need,seoni2023application}, making uncertainty quantification (UQ) a fundamental requirement for safe clinical deployment \cite{lopez2025uncertainty}, with medical image segmentation as a prominent testbed for high-risk applications \cite{huang2024review,li2025evaluation}. Most existing work obtains segmentation uncertainty by sampling an approximate predictive distribution at inference, e.g., via Monte Carlo dropout or deep ensembles \cite{gal2016dropout,lakshminarayanan2017simple}, then processing these samples with measures such as Shannon entropy or voxel-wise variance to produce spatial (aleatoric) uncertainty maps \cite{huang2024review}. In practice, such maps are consumed by visual inspection: a clinician reading them performs an implicit case-level aggregation, dismissing isolated highlighted voxels and attending instead to regions where uncertainty clusters into salient shapes, with a conspicuous cluster prompting closer re-inspection, follow-up, or a second reading. This pooled reading, rather than any single voxel, is the operative signal at the point of care: to inform such decisions, an uncertainty map must flag the cases that experts themselves would flag. Yet uncertainty maps of pixel-to-pixel trained segmentation models are dominated by low-level labeling variability at object boundaries (Fig.\ref{fig:sample}), driven by partial volume effects or minor inter-annotator inconsistencies \cite{yang2023assessing,tohka2014partial}, and thus tend to emphasize disagreement over boundary regions \cite{islam2021spatially,ye2023confidence}. However, segmentation tasks with potentially absent pathological findings involve a semantically distinct and more clinically meaningful ambiguity, which is case-level inter-annotator disagreement about whether a pathology is present at all. For the reader of an uncertainty map, this failure mode is silent: the map remains visually plausible, highlighting boundary regions as expected, while a case in which experts disagree about the very existence of a finding can appear less uncertain than an unambiguous one (Fig.\ref{fig:sample}). Aleatoric uncertainty maps can thus offer false reassurance precisely where caution is warranted. Despite its relevance, this presence ambiguity remains comparatively understudied, a gap we address in this work. Our contributions can be summarized as follows:
\begin{itemize}
    \item We systematically evaluate whether entropy-based aleatoric uncertainty scores capture \emph{existential} ambiguity in lung nodule segmentation (case-level disagreement about the \emph{presence} of a segmentation target) and show that they cannot: entropy-based maps reliably highlight object boundaries (low-level pixel-wise annotator noise) but fail to capture whether a pathology is present at all, regardless of whether uncertainty is aggregated over the ground-truth or the predicted mask, revealing a fundamental mismatch between standard UQ methods and clinically relevant ambiguity.
    \item We show that a lightweight supervised post-hoc ambiguity head, trained on top of frozen segmentation features, recovers case-level annotator disagreement far better than entropy aggregation of MC-dropout and ensemble predictions, and matches or exceeds methods that explicitly model ambiguity under disagreement supervision (Probabilistic 3D-UNet \cite{Kohl2018APU}, Annotator-Confusion 3D-UNet \cite{Tanno2019LearningFN}), consistently across architectures on both \\in-distribution (LIDC) \cite{armato2011lung} and out-of-distribution (LNDb) \cite{pedrosa2019lndb} data.
    \item We offer an explanation of this phenomenon: a qualitative feature-space analysis (UMAP) \cite{McInnes2018UMAPUM} shows that networks trained purely on pixel-wise supervision already encode presence ambiguity in their internal representations.
\end{itemize}
\section{Related Work}
\noindent
\textbf{Bayesian Uncertainty Estimation and Disentanglement}\\
\noindent
Within the supervised learning framework, a probabilistic learner aims to construct a predictive distribution $\hat{p}(\cdot \mid x)$ over outcomes $y \in \mathcal{Y}$ given inputs $x \in \mathcal{X}$. In a Bayesian setting, this corresponds to the posterior predictive distribution, marginalizing over model parameters $\theta$ \cite{murphy2022probabilistic,wimmer2023quantifying}. Predictive uncertainty (PU) is quantified as the Shannon entropy of this distribution:
\begin{equation}
\mathrm{PU}(y \mid x) = H\!\left( \mathbb{E}_{p(\theta \mid \mathcal{D})}[p(y \mid x, \theta)] \right),
\end{equation}
and is commonly decomposed into aleatoric and epistemic components \cite{hullermeier2021aleatoric}:
\begin{equation}
\mathrm{PU}(y \mid x)
\;=\;
\underbrace{\mathbb{E}_{p(\theta \mid \mathcal{D})}[H(p(y \mid x, \theta))]}_{\mathrm{AU}(y \mid x)}
\;+\;
\underbrace{\mathcal{I}(y, \theta \mid x)}_{\mathrm{EU}(y \mid x)}.
\end{equation}
Aleatoric uncertainty (AU) captures irreducible input-dependent ambiguity, such as measurement noise or ambiguous anatomical boundaries, while epistemic uncertainty (EU) reflects reducible lack of knowledge \cite{depeweg2018decomposition}. In practice, AU is approximated via the expected entropy of Monte Carlo dropout samples or deep ensembles \cite{gal2016dropout,lakshminarayanan2017simple}, and its separation from EU (disentanglement) has been studied in general benchmarks \cite{mucsanyi2024benchmarking} and medical image classification \cite{baur2025benchmarking}. However, this decomposition has been called into question: the use of Shannon entropy, conditional entropy, and mutual information as uncertainty proxies, and the assumed additivity of the AU--EU split itself, have been empirically challenged \cite{wimmer2023quantifying,mucsanyi2024benchmarking,baur2025benchmarking}, casting doubt on whether it yields the semantically meaningful separation it is commonly assumed to provide.

\noindent
\textbf{Aleatoric Uncertainty From Spatial Inter-Annotator Disagreement}\\
\noindent
A line of work grounds aleatoric uncertainty directly in inter-annotator disagreement, building on probabilistic segmentation models that augment a U-Net with a conditional latent space to sample globally consistent segmentation hypotheses \cite{Kohl2018APU}. Unlike independent pixel-wise fluctuations, these latent samples correspond to coherent segmentation variants, naturally capturing spatial annotation ambiguity. Subsequent work calibrates this uncertainty by supervising the model against empirical per-pixel label frequencies from multiple annotators via soft-label cross-entropy \cite{Tanno2019LearningFN,hu2019supervised, zhang2020disentangling}. While an important step beyond unsupervised sampling-based UQ, these methods target voxel-level spatial disagreement rather than the case-level existential ambiguity of whether a pathology is present at all, the setting we address here.
\section{Method and Experimental Setup}
\begin{figure}
\includegraphics[width=\textwidth]{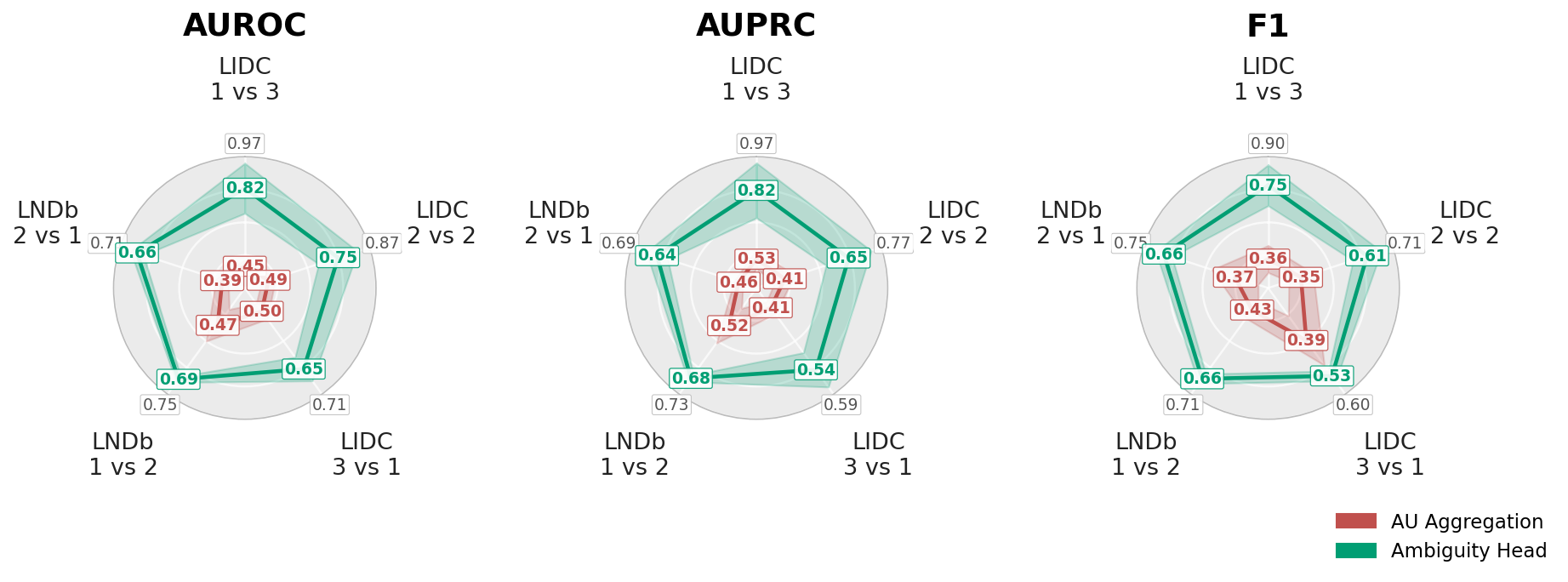}
\caption{Presence ambiguity detection across disagreement levels on LIDC and LNDb. Radar plots compare pixel-wise AU aggregation (mean of Ensemble and MC-Dropout) against post-hoc Ambiguity Head, averaged over four architectures. Entropy-based aggregation stays near-random across all disagreement levels, while the lightweight head recovers a highly discriminative signal from frozen segmentation features.} 
\label{fig:radar}
\end{figure}

We conduct all experiments on the LIDC-IDRI (LIDC) dataset \cite{armato2011lung} (70/15/15 train/val/test split), a large-scale thoracic CT collection with a two-stage annotation protocol in which radiologists first labeled independently and then revised after reviewing their peers' assessments, so that remaining disagreements reflect genuine clinical ambiguity about nodule presence rather than incidental labeling noise. For generalization, we additionally evaluate on the LNDb dataset \cite{pedrosa2019lndb} as an external validation cohort of completely unseen data, with no retraining or supervision.

\subsubsection{Case-Level Ambiguity Formulation}
The LIDC and LNDb datasets provide up to $K$ independent radiologist annotations per scan ($K=4$ and  $K=3$, respectively). Each annotator supplies a binary segmentation mask $m_k \in \{0,1\}^{H \times W \times D}$ indicating the presence of a lung nodule. We define a binary presence indicator for each annotator:
\begin{equation}
\mathbf{1}_k(x) =
\begin{cases}
1 & \text{if } |m_k| > 0 \\
0 & \text{otherwise,}
\end{cases}
\end{equation}
where $|m_k|$ denotes the number of foreground voxels in mask $m_k$. The empirical presence agreement for a case $x$ is then:
\begin{equation}
A(x) = \frac{1}{K} \sum_{k=1}^{K} \mathbf{1}_k(x) \in [0,1].
\end{equation}
We define \emph{non-ambiguous cases} as those with unanimous agreement on presence or absence, $A(x) \in \{0, 1\}$, and \emph{ambiguous cases} as those in which annotators disagree, $0 < A(x) < 1$. Ambiguity detection is then formulated as a binary classification problem:
\begin{equation}
y_{\text{amb}}(x) =
\begin{cases}
1 & \text{if } 0 < A(x) < 1 \\
0 & \text{otherwise.}
\end{cases}
\end{equation}
The resulting label distinguishes global presence ambiguity from pixel-wise boundary disagreement, covering different partial-disagreement configurations (1-vs-3, 2-vs-2, and 3-vs-1 for LIDC; 1-vs-2 and 2-vs-1 for LNDb).

\subsubsection{Case-Level Uncertainty Aggregation}
Since ambiguity is defined at the case level, voxel-wise uncertainty maps
must be aggregated into a scalar score. As reference region for the aggregation we use the union of the annotator
masks, $M(x) = \{ v_i : \max_k m_k(v_i) = 1 \}$, i.e., a voxel belongs to
$M(x)$ if at least one annotator labeled it as foreground. $M(x)$ is
therefore well-defined under annotator disagreement and non-empty for every
evaluated case, as each case contains at least one non-empty annotation.
We provide results for aggregating over the ground truth mask as well as the predicted mask. When aggregating over the predicted mask instead,
$\hat{M}(x) = \{ v_i : \hat{y}_{v_i} = 1 \}$, the prediction may be empty
(roughly 5\% of cases); these cases are excluded from the predicted-mask
evaluation. We aggregate aleatoric uncertainty by computing the mean of
$\mathrm{AU}(v_i)$ over all voxels within the ground-truth mask:
\begin{equation}
U_{\mathrm{AU}}(x) = \frac{1}{|M(x)|} \sum_{v_i \in M(x)} \mathrm{AU}(v_i).
\end{equation}
In our experiments we evaluate how well the aggregated uncertainty score
$U_{\mathrm{AU}}(x)$ aligns with the case-level ambiguity label
$y_{\text{amb}}(x)$.

\subsubsection{Post-hoc Ambiguity Head}
Our post-hoc ambiguity head trained on frozen segmentation features is defined as follows: let $f_\theta(x)$ denote the feature representation extracted at the architecture bottleneck (the most compressed latent representation, prior to any decoder upsampling) of the frozen, pretrained segmentation backbone. Operating on this representation, we attach a trainable head $g_\phi$ consisting of attention pooling followed by a 3-layer MLP with ReLU activations and a sigmoid output. We optimize:

\begin{equation}
\phi^* = \arg\min_\phi \; \mathcal{L}_{\text{BCE}}\!\left(g_\phi(f_\theta(x)),\, y_{\text{amb}}(x)\right),
\end{equation}
where $\mathcal{L}_{\text{BCE}}$ denotes binary cross-entropy loss. 
The resulting scalar output \newline $U_{\mathrm{amb}}(x) = g_{\phi^*}(f_\theta(x))$ serves as ambiguity estimate akin to $U_{\mathrm{AU}}(x)$.

\begin{figure}[!h]
\includegraphics[width=\textwidth]{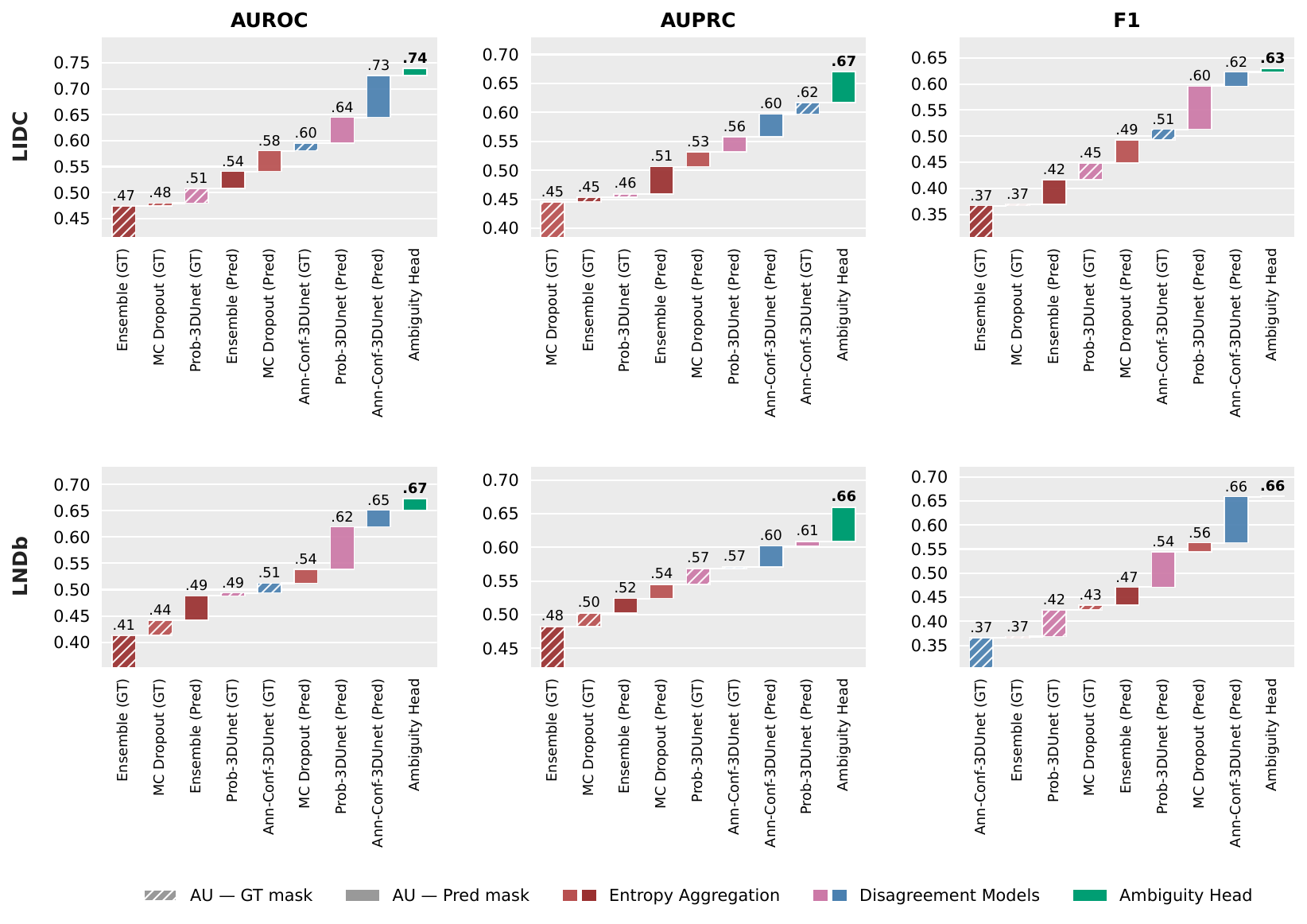}
\caption{Sorted performance metrics, averaged across model architectures and
ambiguity thresholds. Aleatoric uncertainty (AU) baselines appear twice,
differing in the region over which voxel-wise uncertainty is aggregated into
a score: the ground-truth mask (hatched) or the predicted mask (solid).
Using the predicted mask consistently improves both the unsupervised
entropy-aggregation methods and the supervised disagreement-modelling
baselines (Prob-3DUnet, Ann-Conf-3DUnet). The Ambiguity Head surpasses all
baseline variants across metrics in almost all cases, except for Ann-Conf-3DUnet (prediction mask aggregation) where it performs on par. Notably, this performance also holds in the external validation cohort (LNDb). Standard deviations are omitted here, as averaging over disagreement levels,
models, and runs renders them uninformative. Non-aggregated metrics and
standard deviations per disagreement level and model are reported in
Appendix~A.2.}
\label{fig:waterfall}
\end{figure}

\subsubsection{Experimental Details}

Details regarding experimental setting, training and evaluation beyond what is stated above are provided in appendix A.1.

\section{Results}

\begin{figure}[!h]
\includegraphics[width=\textwidth]{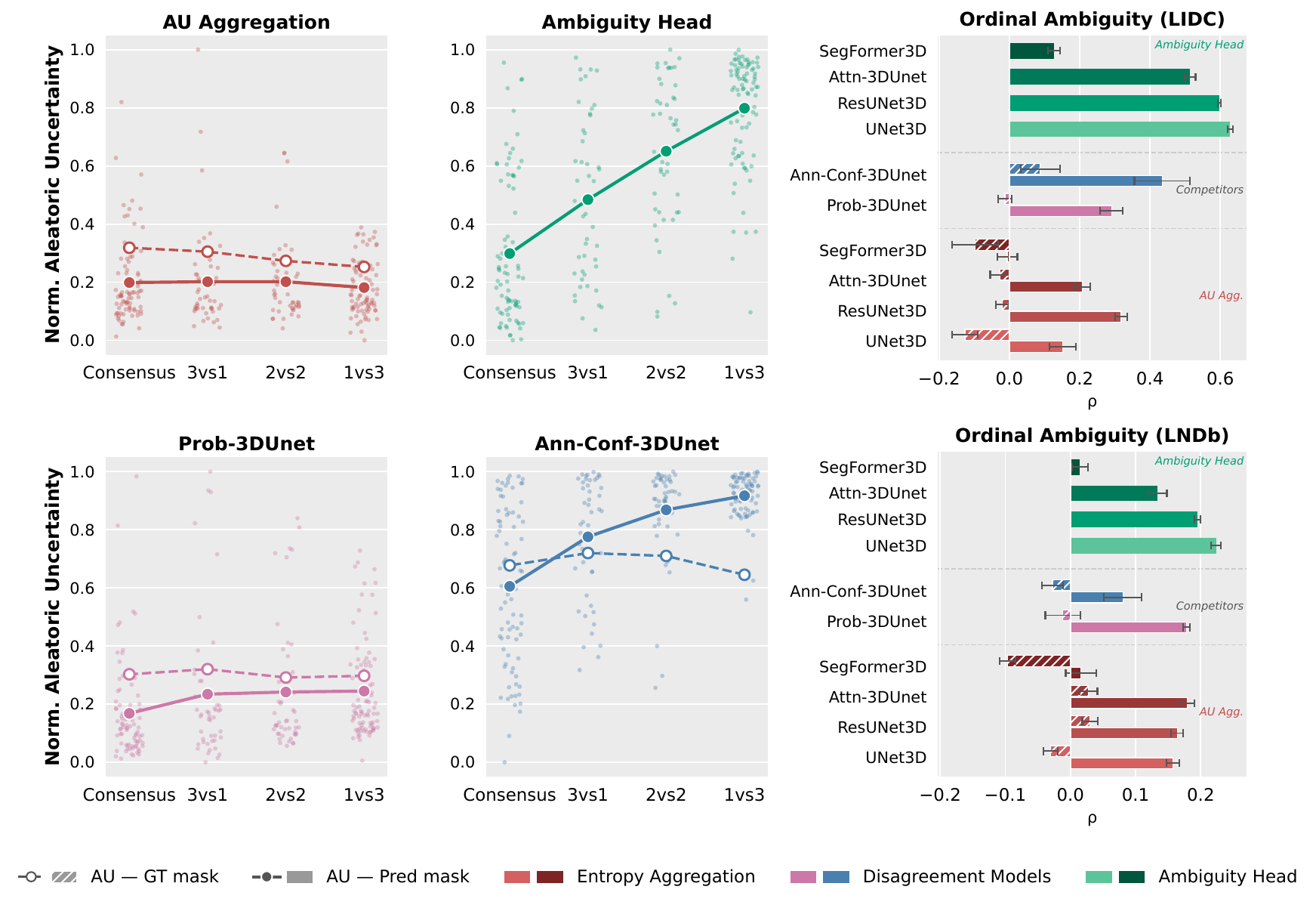}
\caption{Ordinal correlation between uncertainty scores and ambiguity severity. Scatter panels (left, mid) show an exemplary distribution of uncertainty scores (UNet-3D). The summary (right) reports Spearman correlations ($\rho$) across both cohorts, all backbones, and all methods. The Ambiguity Head stratifies the full spectrum of disagreement severity rather than only binary ambiguity, whereas pixel-wise AU aggregation and the supervised baselines track severity only weakly or inconsistently.} 
\label{fig:scatter}
\end{figure}

The radar plots in Figure~\ref{fig:radar} compare entropy aggregation (mean of MC-dropout and ensemble, which behave near-identically) against the ambiguity head, averaged over all four architectures and broken down by disagreement configuration. Across both cohorts, entropy aggregation stays close to random at every disagreement level, whereas the ambiguity head is strongly discriminative throughout. Figure~\ref{fig:waterfall} aggregates over models and disagreement levels, adds the supervised competitors, and reports every baseline for both aggregation regions: the ambiguity head reaches $0.74$ AUROC, $0.67$ AUPRC, and $0.63$ F1 on LIDC, against at most $0.48$ AUROC for entropy aggregation over the ground-truth mask. Aggregating over the predicted mask instead consistently improves all baselines: entropy aggregation rises to at most $0.58$ AUROC, remaining far below the ambiguity head, while the methods that explicitly model ambiguity under disagreement supervision benefit most, with the Annotator-Confusion 3D-UNet improving from $0.60$ to $0.73$ AUROC and the Probabilistic 3D-UNet from $0.51$ to $0.64$. The ambiguity head thus remains the strongest method across all metrics, though the margin over the best supervised competitor narrows under predicted-mask aggregation, down to on-par performance on F1 ($0.63$ vs.\ $0.62$). The same ordering holds on LNDb: $0.67$ AUROC for the ambiguity head against at most $0.54$ for entropy aggregation and $0.65$ for the Annotator-Confusion 3D-UNet, with both reaching $0.66$ F1. Beyond binary detection, Figure~\ref{fig:scatter} measures whether each score ranks cases by disagreement severity via the Spearman correlation with the ordinal agreement tier. The ambiguity head increases monotonically across tiers and attains the strongest correlation across both cohorts and all backbones. Entropy-based AU aggregation over the ground-truth mask is near-zero and flat; switching to the predicted mask recovers a moderate correlation (up to $\rho = 0.43$ for the Annotator-Confusion 3D-UNet on LIDC), which nevertheless stays well below the ambiguity head ($0.63$), while on LNDb all correlations are compressed and the ambiguity head retains a smaller lead ($0.22$ vs.\ at most $0.18$). The ambiguity head's discriminative power scales modestly with segmentation quality, though this is confounded with backbone family, as the three UNet variants yield more informative heads than the transformer-based SegFormer3D, hinting that the optimal feature-extraction point may be architecture-dependent. All backbones reach segmentation performance in line with the literature \cite{xi2025coreformer,zhang2020disentangling}: 0.80--0.84 Dice for the UNet variants, 0.78 for SegFormer3D, and 0.70 for Annotator-Confusion 3D-UNet. Per-architecture and per-disagreement level metrics for both aggregation regions are reported in Appendix~A.2. Finally, the UMAP projection in Figure~\ref{fig:umaps} offers insight into why the ambiguity head succeeds: consensus and ambiguous cases form clearly separable clusters across architectures, indicating that presence ambiguity is already encoded in the encoder representations of pixel-wise trained networks.

\begin{figure}
\includegraphics[width=\textwidth]{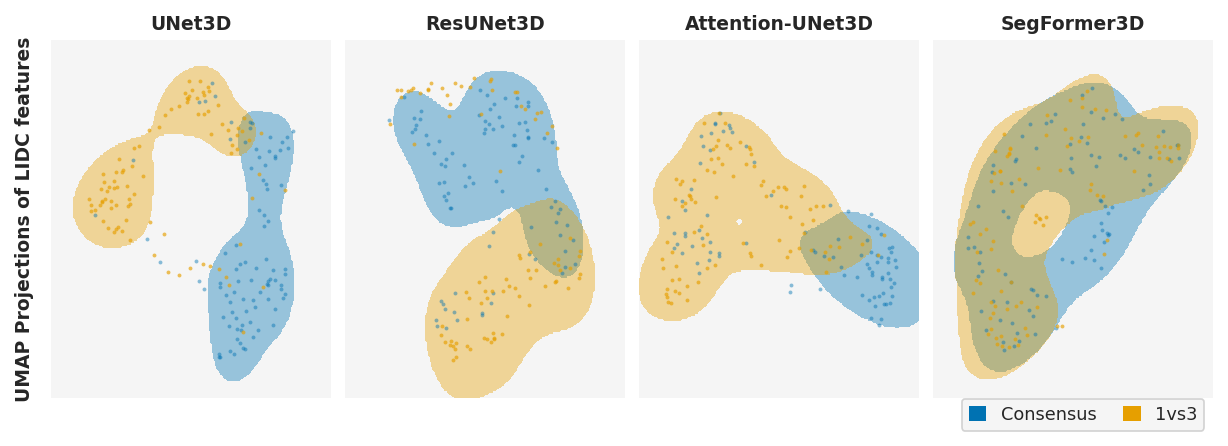}
\caption{UMAP projection of frozen model features (1vs3). The clustering indicates presence ambiguity is already well-represented in the latent space, i.e. Ambiguity Head does not learn from scratch but recovers a signal the standard pipeline fails to transport.}
\label{fig:umaps}
\end{figure}

\section{Conclusion}
Across four segmentation architectures and two datasets (one an external
validation cohort), we showed that entropy-based aleatoric uncertainty is
ill-suited for detecting case-level presence ambiguity for lung nodule
segmentation: under standard training, spatial entropy maps capture boundary
noise and minor drawing variation but carry near-zero signal for whether a
pathology is present at all. This gap between the theoretical promise of
aleatoric uncertainty and its practical behavior is critical for clinical
deployment, where misleading estimates can directly influence diagnostic
decisions, and documenting it is the central contribution of this work. The
ambiguity head should be read in this light: not as a proposed competitor that improves ambiguity detection over baseline methods, 
but as a deliberately supervision-advantaged upper bound that makes the
failure quantifiable. It demonstrates that presence ambiguity is readily
recoverable from the frozen features of pixel-wise trained networks (i.e. the
signal being present, but discarded by entropy aggregation on the pixel-wise segmentation output) and that even methods trained with explicit disagreement supervision recover it only partially. We therefore recommend against using
entropy-based uncertainty as a proxy for clinical ambiguity when supervision
ground truths are available. As this work mainly aims to highlight the
underexplored problem of presence ambiguity, the ambiguity head is only a
simple proof of concept; future work should explore richer ways to
incorporate presence ambiguity during training and extend the analysis to
other structures and modalities, even though case-level ambiguity with
multiple annotator labels remains rare in public datasets, itself a
limitation the community should address.

\subsubsection{Acknowledgments.}
This work was supported by the Senate of Berlin and the European Commision's Digital Europe Programme (DIGITAL) as grant TEF-Health (101100700).

\subsubsection{Disclosure of Interests.}
The authors have no competing interests to declare that are relevant to the content of this article.

\bibliographystyle{splncs04}
\bibliography{references}
\newpage
\section*{Appendix}

\subsubsection*{A.1}

We evaluate all methods on their ability to discriminate ambiguous from non-ambiguous cases using AUROC, AUPRC, and F1 score, with all test cases held out from segmentation training and ambiguity post-hoc training of the head. For LIDC, the test set comprises 88 (1vs3), 47 (2vs2), 47(3vs1) and 80 non-ambiguous cases; the optimal F1 threshold is selected on the validation set. For LNDb, we evaluate on a class-balanced subset obtained by down-sampling each class to the minority-class size. We report the mean over 5 differently initialized runs. All models were trained for 100 epochs with early stopping of 15 on best val dice score, using gt masks with a minimum consensus of 2 and 0.5 threshold for foreground prediction. Ambiguity head post-hoc training is done for 10 epochs. The per-voxel aleatoric uncertainty $\mathrm{AU}(v_i)$ is derived via stochastic sampling for MC-dropout models and Probabilistic 3D-UNet ($S{=}100$ samples, dropout=0.1), across $N{=}3$ members for the ensemble, and across $K$ per-annotator predictions $p_k(v_i)$ obtained from learned annotator-specific confusion matrices for the Annotator-Confusion 3D-UNet, where $\mathrm{AU}(v_i) = \frac{1}{K}\sum_{k=1}^{K} \mathcal{H}(p_k(v_i))$ and $\mathcal{H}(\cdot)$ denotes the entropy.

\subsubsection*{A.2}

In Tables 1 and 2 below we provide detailed metrics underlying the reported metrics in Figures \ref{fig:radar} and \ref{fig:waterfall}. Tables 3 and 4 provide results as in Figures \ref{fig:waterfall} and \ref{fig:scatter} where $U_{\mathrm{AU}}(x)$ was aggregated over the prediction mask instead of the ground truth mask.

\begin{table}[!h]
\centering
\caption{Presence ambiguity detection performance on disagreement level.}
\label{tab:ambiguity_results}
\setlength{\tabcolsep}{3pt} 
\footnotesize 
\resizebox{\textwidth}{!}{
\begin{tabular}{ll ccc ccc ccc}
\toprule
& & \multicolumn{3}{c}{\textbf{AU (Ensemble)}} & \multicolumn{3}{c}{\textbf{AU (MC-Dropout)}} & \multicolumn{3}{c}{\textbf{Ambiguity Head}} \\
\cmidrule(lr){3-5} \cmidrule(lr){6-8} \cmidrule(lr){9-11}
\textbf{Split} & \textbf{Model} & \textbf{AUROC} & \textbf{AUPRC} & \textbf{F1} & \textbf{AUROC} & \textbf{AUPRC} & \textbf{F1} & \textbf{AUROC} & \textbf{AUPRC} & \textbf{F1} \\
\midrule
\multirow{4}{*}{LIDC 1vs3} 
& Attn-3DUnet & 0.45$\pm$0.02 & 0.53$\pm$0.02 & 0.40$\pm$0.03 & 0.48$\pm$0.04 & 0.54$\pm$0.02 & 0.38$\pm$0.09 & \textbf{0.86}$\pm$0.04 & \textbf{0.87}$\pm$0.03 & \textbf{0.78}$\pm$0.05 \\
& 3D-ResUnet   & 0.48$\pm$0.02 & 0.57$\pm$0.02 & 0.44$\pm$0.03 & 0.49$\pm$0.03 & 0.57$\pm$0.03 & 0.47$\pm$0.04 & \textbf{0.90}$\pm$0.01 & \textbf{0.89}$\pm$0.03 & \textbf{0.82}$\pm$0.04 \\
& SegFormer3D & 0.43$\pm$0.04 & 0.58$\pm$0.04 & 0.32$\pm$0.09 & 0.42$\pm$0.12 & 0.53$\pm$0.08 & 0.23$\pm$0.17 & \textbf{0.61}$\pm$0.04 & \textbf{0.62}$\pm$0.03 & \textbf{0.57}$\pm$0.04 \\
& 3D-Unet      & 0.39$\pm$0.02 & 0.45$\pm$0.01 & 0.32$\pm$0.04 & 0.42$\pm$0.07 & 0.47$\pm$0.03 & 0.31$\pm$0.17 & \textbf{0.91}$\pm$0.02 & \textbf{0.92}$\pm$0.01 & \textbf{0.83}$\pm$0.02 \\
\midrule
\multirow{4}{*}{LIDC 2vs2} 
& Attn-3DUnet & 0.52$\pm$0.01 & 0.42$\pm$0.01 & 0.38$\pm$0.09 & 0.52$\pm$0.03 & 0.40$\pm$0.01 & 0.41$\pm$0.09 & \textbf{0.76}$\pm$0.04 & \textbf{0.66}$\pm$0.04 & \textbf{0.60}$\pm$0.06 \\
& 3D-ResUnet   & 0.50$\pm$0.01 & 0.45$\pm$0.02 & 0.39$\pm$0.04 & 0.50$\pm$0.02 & 0.43$\pm$0.02 & 0.38$\pm$0.03 & \textbf{0.82}$\pm$0.03 & \textbf{0.72}$\pm$0.04 & \textbf{0.67}$\pm$0.06 \\
& SegFormer3D & 0.50$\pm$0.03 & 0.46$\pm$0.06 & 0.27$\pm$0.05 & 0.48$\pm$0.07 & 0.42$\pm$0.06 & 0.28$\pm$0.14 & \textbf{0.63}$\pm$0.03 & \textbf{0.51}$\pm$0.02 & \textbf{0.53}$\pm$0.04 \\
& 3D-Unet      & 0.43$\pm$0.02 & 0.34$\pm$0.01 & 0.36$\pm$0.09 & 0.46$\pm$0.05 & 0.36$\pm$0.04 & 0.38$\pm$0.06 & \textbf{0.79}$\pm$0.03 & \textbf{0.70}$\pm$0.03 & \textbf{0.66}$\pm$0.05 \\
\midrule
\multirow{4}{*}{LIDC 3vs1} 
& Attn-3DUnet & 0.51$\pm$0.02 & 0.43$\pm$0.02 & 0.45$\pm$0.05 & 0.51$\pm$0.02 & 0.41$\pm$0.02 & 0.43$\pm$0.08 & \textbf{0.66}$\pm$0.06 & \textbf{0.57}$\pm$0.05 & \textbf{0.52}$\pm$0.08 \\
& 3D-ResUnet   & 0.52$\pm$0.03 & 0.41$\pm$0.02 & 0.43$\pm$0.01 & 0.51$\pm$0.03 & 0.40$\pm$0.03 & 0.43$\pm$0.05 & \textbf{0.65}$\pm$0.02 & \textbf{0.54}$\pm$0.03 & \textbf{0.53}$\pm$0.02 \\
& SegFormer3D & 0.46$\pm$0.04 & 0.42$\pm$0.05 & 0.20$\pm$0.05 & 0.47$\pm$0.07 & 0.40$\pm$0.05 & 0.26$\pm$0.22 & \textbf{0.60}$\pm$0.02 & \textbf{0.48}$\pm$0.03 & \textbf{0.51}$\pm$0.04 \\
& 3D-Unet      & 0.50$\pm$0.01 & 0.39$\pm$0.01 & 0.45$\pm$0.08 & 0.51$\pm$0.04 & 0.39$\pm$0.02 & 0.48$\pm$0.08 & \textbf{0.69}$\pm$0.02 & \textbf{0.57}$\pm$0.01 & \textbf{0.56}$\pm$0.03 \\
\midrule
\multirow{4}{*}{LNDb 1vs2} 
& Attn-3DUnet & 0.53$\pm$0.03 & 0.59$\pm$0.06 & 0.46$\pm$0.07 & 0.53$\pm$0.04 & 0.58$\pm$0.06 & 0.49$\pm$0.07 & \textbf{0.69}$\pm$0.02 & \textbf{0.68}$\pm$0.04 & \textbf{0.64}$\pm$0.10 \\
& 3D-ResUnet   & 0.48$\pm$0.06 & 0.53$\pm$0.05 & 0.37$\pm$0.14 & 0.52$\pm$0.05 & 0.55$\pm$0.06 & 0.51$\pm$0.10 & \textbf{0.69}$\pm$0.04 & \textbf{0.67}$\pm$0.04 & \textbf{0.68}$\pm$0.04 \\
& SegFormer3D & 0.34$\pm$0.04 & 0.42$\pm$0.04 & 0.40$\pm$0.37 & 0.38$\pm$0.06 & 0.45$\pm$0.04 & 0.39$\pm$0.32 & \textbf{0.67}$\pm$0.05 & \textbf{0.67}$\pm$0.06 & \textbf{0.64}$\pm$0.11 \\
& 3D-Unet      & 0.47$\pm$0.03 & 0.50$\pm$0.02 & 0.39$\pm$0.05 & 0.51$\pm$0.05 & 0.55$\pm$0.04 & 0.47$\pm$0.07 & \textbf{0.71}$\pm$0.01 & \textbf{0.70}$\pm$0.04 & \textbf{0.68}$\pm$0.04 \\
\midrule
\multirow{4}{*}{LNDb 2vs1} 
& Attn-3DUnet & 0.34$\pm$0.02 & 0.45$\pm$0.03 & 0.09$\pm$0.10 & 0.36$\pm$0.03 & 0.45$\pm$0.03 & 0.48$\pm$0.27 & \textbf{0.67}$\pm$0.02 & \textbf{0.65}$\pm$0.04 & \textbf{0.69}$\pm$0.05 \\
& 3D-ResUnet   & 0.40$\pm$0.06 & 0.45$\pm$0.04 & 0.45$\pm$0.29 & 0.40$\pm$0.04 & 0.47$\pm$0.03 & 0.58$\pm$0.16 & \textbf{0.62}$\pm$0.04 & \textbf{0.60}$\pm$0.04 & \textbf{0.62}$\pm$0.06 \\
& SegFormer3D & 0.37$\pm$0.04 & 0.44$\pm$0.02 & 0.48$\pm$0.27 & 0.42$\pm$0.05 & 0.49$\pm$0.03 & 0.29$\pm$0.23 & \textbf{0.67}$\pm$0.01 & \textbf{0.67}$\pm$0.03 & \textbf{0.64}$\pm$0.05 \\
& 3D-Unet      & 0.38$\pm$0.01 & 0.47$\pm$0.01 & 0.31$\pm$0.20 & 0.41$\pm$0.03 & 0.48$\pm$0.03 & 0.25$\pm$0.14 & \textbf{0.66}$\pm$0.01 & \textbf{0.64}$\pm$0.03 & \textbf{0.70}$\pm$0.04 \\
\bottomrule
\end{tabular}%
}
\label{tab:entropy_agg}
\end{table}

\begin{table}[!h]
\centering
\caption{Presence ambiguity detection performance compared with supervised competitors.}
\label{tab:ambiguity_baselines_comparison}
\scriptsize 
\renewcommand{\arraystretch}{0.95} 
\begin{tabular*}{\textwidth}{@{\extracolsep{\fill}}ll ccc}
\toprule
\textbf{Split} & \textbf{Model} & \textbf{AUROC} & \textbf{AUPRC} & \textbf{F1} \\
\midrule
\multirow{3}{*}{LIDC 1vs3} 
& \textbf{Ambiguity Head (mean)} & \textbf{0.82}$\pm$0.14 & \textbf{0.82}$\pm$0.14 & \textbf{0.75}$\pm$0.12 \\
& Ann-Conf-3DUnet           & 0.54$\pm$0.09          & 0.68$\pm$0.09          & 0.55$\pm$0.08          \\
& Prob-3DUnet                    & 0.49$\pm$0.03          & 0.56$\pm$0.01          & 0.48$\pm$0.10          \\
\midrule
\multirow{3}{*}{LIDC 2vs2} 
& \textbf{Ambiguity Head (mean)} & \textbf{0.75}$\pm$0.08 & \textbf{0.65}$\pm$0.10 & \textbf{0.61}$\pm$0.06 \\
& Ann-Conf-3DUnet           & 0.63$\pm$0.12          & 0.63$\pm$0.14          & 0.51$\pm$0.13          \\
& Prob-3DUnet                    & 0.48$\pm$0.01          & 0.40$\pm$0.02          & 0.40$\pm$0.03          \\
\midrule
\multirow{3}{*}{LIDC 3vs1} 
& \textbf{Ambiguity Head (mean)} & \textbf{0.65}$\pm$0.04 & \textbf{0.54}$\pm$0.04 & \textbf{0.53}$\pm$0.02 \\
& Ann-Conf-3DUnet           & 0.61$\pm$0.09          & \textbf{0.54}$\pm$0.10 & 0.47$\pm$0.16          \\
& Prob-3DUnet                    & 0.55$\pm$0.04          & 0.43$\pm$0.02          & 0.47$\pm$0.06          \\
\midrule
\multirow{3}{*}{LNDb 1vs2} 
& \textbf{Ambiguity Head (mean)} & \textbf{0.69}$\pm$0.01 & \textbf{0.68}$\pm$0.01 & \textbf{0.66}$\pm$0.02 \\
& Ann-Conf-3DUnet           & 0.50$\pm$0.08          & 0.57$\pm$0.05          & 0.35$\pm$0.18          \\
& Prob-3DUnet                    & 0.55$\pm$0.06          & 0.60$\pm$0.07          & 0.46$\pm$0.04          \\
\midrule
\multirow{3}{*}{LNDb 2vs1} 
& \textbf{Ambiguity Head (mean)} & \textbf{0.66}$\pm$0.02 & \textbf{0.64}$\pm$0.03 & \textbf{0.66}$\pm$0.04 \\
& Ann-Conf-3DUnet           & 0.52$\pm$0.04          & 0.57$\pm$0.02          & 0.38$\pm$0.14          \\
& Prob-3DUnet                    & 0.44$\pm$0.07          & 0.53$\pm$0.07          & 0.39$\pm$0.26          \\
\bottomrule
\end{tabular*}
\label{tab:competitors}
\end{table}

Below we provide results as in Figures 3 and 4 aggregated on prediction mask instead of ground truth mask.  

\begin{table}[t]
\centering
\caption{Presence ambiguity detection performance  across architectures and disagreement thresholds, on LIDC and the external LNDb cohort for aggregation of $U_{\mathrm{AU}}(x)$ on the prediction mask instead of ground truth mask.}
\label{tab:aggregate}
\scriptsize 
\renewcommand{\arraystretch}{0.95} 
\begin{tabular}{lcccccc}
\toprule
& \multicolumn{3}{c}{LIDC} & \multicolumn{3}{c}{LNDb} \\
\cmidrule(lr){2-4} \cmidrule(lr){5-7}
Method & AUROC & AUPRC & F1 & AUROC & AUPRC & F1 \\
\midrule
Ambiguity Head      & \textbf{0.74} & \textbf{0.67} & \textbf{0.63} & \textbf{0.67} & \textbf{0.66} & \textbf{0.66} \\
Ann-Conf-3DUnet & 0.73 & 0.60 & 0.62 & 0.65 & 0.60 & 0.66 \\
Prob-3DUnet           & 0.64 & 0.56 & 0.60 & 0.62 & 0.61 & 0.55 \\
MC-Dropout          & 0.58 & 0.53 & 0.49 & 0.54 & 0.55 & 0.56 \\
Ensemble            & 0.54 & 0.51 & 0.42 & 0.49 & 0.52 & 0.47 \\
\bottomrule
\end{tabular}
\end{table}

\begin{table}[t]
\centering
\caption{Ordinal correlation (Spearman $\rho$) between uncertainty scores and disagreement severity for aggregation of $U_{\mathrm{AU}}(x)$ on the prediction mask instead of ground truth mask.}
\scriptsize 
\renewcommand{\arraystretch}{0.95} 
\label{tab:ordinal}
\begin{tabular}{llcc}
\toprule
Method & Backbone & LIDC $\rho$ & LNDb $\rho$ \\
\midrule
\multirow{6}{*}{AU Aggregation}
 & Ann. Confusion & 0.43\,$\pm$\,0.08 & 0.08\,$\pm$\,0.03 \\
 & 3D-ResUnet     & 0.32\,$\pm$\,0.02 & 0.16\,$\pm$\,0.01 \\
 & Prob-UNet     & 0.29\,$\pm$\,0.03 & 0.18\,$\pm$\,0.01 \\
 & Attn-3DUnet    & 0.21\,$\pm$\,0.02 & 0.18\,$\pm$\,0.01 \\
 & 3D-Unet        & 0.15\,$\pm$\,0.04 & 0.16\,$\pm$\,0.01 \\
 & SegFormer3D   & $-0.01$\,$\pm$\,0.03 & 0.02\,$\pm$\,0.02 \\
\midrule
\multirow{4}{*}{Ambiguity Head}
 & 3D-Unet        & \textbf{0.63\,$\pm$\,0.01} & \textbf{0.22\,$\pm$\,0.01} \\
 & 3D-ResUnet     & 0.60\,$\pm$\,0.00 & 0.20\,$\pm$\,0.01 \\
 & Attn-3DUnet    & 0.51\,$\pm$\,0.02 & 0.13\,$\pm$\,0.02 \\
 & SegFormer3D   & 0.13\,$\pm$\,0.02 & 0.02\,$\pm$\,0.01 \\
\bottomrule
\end{tabular}
\end{table}

\end{document}